%% file: wsdm2020.tex
\documentclass[sigconf]{acmart}

\usepackage[utf8]{inputenc}
\usepackage{graphicx}
\usepackage{float}
\usepackage{subfigure}
\usepackage{algorithm}
\usepackage{algorithmic}
\usepackage{xcolor}
\usepackage{booktabs} 
\usepackage{mathtools}
\usepackage{amsmath,bm}
\usepackage{makecell}
\usepackage{multirow}
\usepackage{multicol}
\usepackage{array}
\usepackage{comment}
\usepackage{amsfonts,amssymb}
\usepackage{textcomp}
\usepackage{fmtcount}
\usepackage{flushend}

\usepackage{balance}

\newcommand{\customcomment}[3]{\textcolor{#1}{[#2:#3]}}
\newcommand{\chj}[1]{\customcomment{purple}{CHJ}{#1}}

\newcommand{\zny}[1]{\customcomment{blue}{ZNY}{#1}} 

\newcommand{\todo}[1]{\customcomment{red}{TODO}{#1}}

\AtBeginDocument{%
  \providecommand\BibTeX{{%
    \normalfont B\kern-0.5em{\scshape i\kern-0.25em b}\kern-0.8em\TeX}}}

\copyrightyear{2020} 
\acmYear{2020} 
\setcopyright{acmcopyright}
\acmConference[WSDM '20]{The Thirteenth ACM International Conference on Web Search and Data Mining}{February 3--7, 2020}{Houston, TX, USA}
\acmBooktitle{The Thirteenth ACM International Conference on Web Search and Data Mining (WSDM '20), February 3--7, 2020, Houston, TX, USA}
\acmPrice{15.00}
\acmDOI{10.1145/3336191.3371796}
\acmISBN{978-1-4503-6822-3/20/02}

\acmSubmissionID{wsdm305}

\begin{document}

\title{Meta-Learning with Dynamic-Memory-Based Prototypical Network for Few-Shot Event Detection}

\author{Shumin Deng}
\affiliation{
  \institution{Zhejiang University}
  \country{China}
}
\email{231sm@zju.edu.cn}

\author{Ningyu Zhang}
\affiliation{
  \institution{Alibaba Group}
  \country{China}
}
\email{ningyu.zny@alibaba-inc.com}

\author{Jiaojian Kang}
\affiliation{
  \institution{Zhejiang University}
  \country{China}
}
\email{kangjiaojian@zju.edu.cn}

\author{Yichi Zhang}
\affiliation{
  \institution{Alibaba Group}
  \country{China}
}
\email{yichi.zyc@alibaba-inc.com}

\author{Wei Zhang} 
\affiliation{
  \institution{Alibaba Group}
  \country{China}
}
\email{lantu.zw@alibaba-inc.com}

\author{Huajun Chen}
\affiliation{
  \institution{Zhejiang University}
  \country{China}
}
\email{huajunsir@zju.edu.cn}



\renewcommand{\shortauthors}{Deng et al.}

\begin{abstract}
Event detection (ED), a sub-task of event extraction, involves identifying triggers and categorizing event mentions.
Existing methods primarily rely upon supervised learning and require large-scale labeled event datasets which are unfortunately not readily available in many real-life applications. In this paper, we consider and reformulate the ED task with limited labeled data as a \emph{Few-Shot Learning} problem. We propose a \emph{Dynamic-Memory-Based Prototypical Network (DMB-PN)}, which exploits \emph{Dynamic Memory Network (DMN)} to not only learn better prototypes for event types, but also produce more robust sentence encodings for event mentions. Differing from vanilla prototypical networks simply computing event prototypes by averaging, which only consume event mentions once, our model is more robust and is capable of distilling contextual information from event mentions for multiple times due to the multi-hop mechanism of DMNs. The experiments show that DMB-PN not only deals with sample scarcity better than a series of baseline models but also performs more robustly when the variety of event types is relatively large and the instance quantity is extremely small.
\end{abstract}



\maketitle

\input{introduction}

\input{related_work}

\input{method}

\input{experiment}

\input{conclusion}

\section{Acknowledgements}
We want to express gratitude to the anonymous reviewers for their hard work and kind
comments, which will further improve our work in the future. 
This work is funded by 
NSFC 91846204/61473260, 
national key research program YS2018YFB140004, 
and Alibaba CangJingGe(Knowledge Engine) Research Plan.

\clearpage
\bigskip

\balance

\bibliographystyle{ACM-Reference-Format}
\bibliography{wsdm2020}


\end{document}

%% file: introduction.tex
\section{Introduction} \label{sec:intro}
Event extraction (EE) is a task aimed at extracting structural event information from unstructured texts. An event is defined as a specific occurrence involving participants, described in an \emph{event mention} \cite{chen2015event}. The main word or nugget (typically a verb or a noun) that most clearly expresses the occurrence of an event is called a \emph{trigger} \cite{chen2015event}.
In this paper, we focus on the event detection (ED) task, a subtask of EE, which aims to locate the triggers of specified event types in texts.
For example, in the sentence ‘‘He is \emph{married} to the Iraqi microbiologist known as Dr. Germ.'', the ED task should detect the word ‘\emph{married}' as a trigger for the event type ‘\emph{Marry}'.

Typical approaches for ED follow a supervised learning paradigm, which typically relies upon large sets of labeled data and they are unfortunately not readily available in many real-life applications. Even for widely-used ACE-2005 corpus, about $25\%$ event types have less than $20$ intances \cite{nguyen2016two}. 
More importantly, new event types tend to emerge frequently in practice, whereas most traditional models are hardly able to classify new events correctly if only a small number of samples for these new event types are given.

Intuitively, people can promptly assimilate new knowledge and deduce new classes by learning from few instances, due to the human brain's ability to synthesis, adapt and transfer knowledge from different learned classes, which is known as the ability of ‘‘learning to learn'' or ‘‘meta-learning'' \cite{finn2017model,santoro2016meta,snell2017prototypical}. The process of developing a classifier which must generalize to new classes from only a small number of samples at a rapid pace is also commonly referred as \emph{few-shot learning (FSL)} \cite{snell2017prototypical}.

\begin{figure}[!hpt]
  \centering
  \includegraphics[scale=0.31]{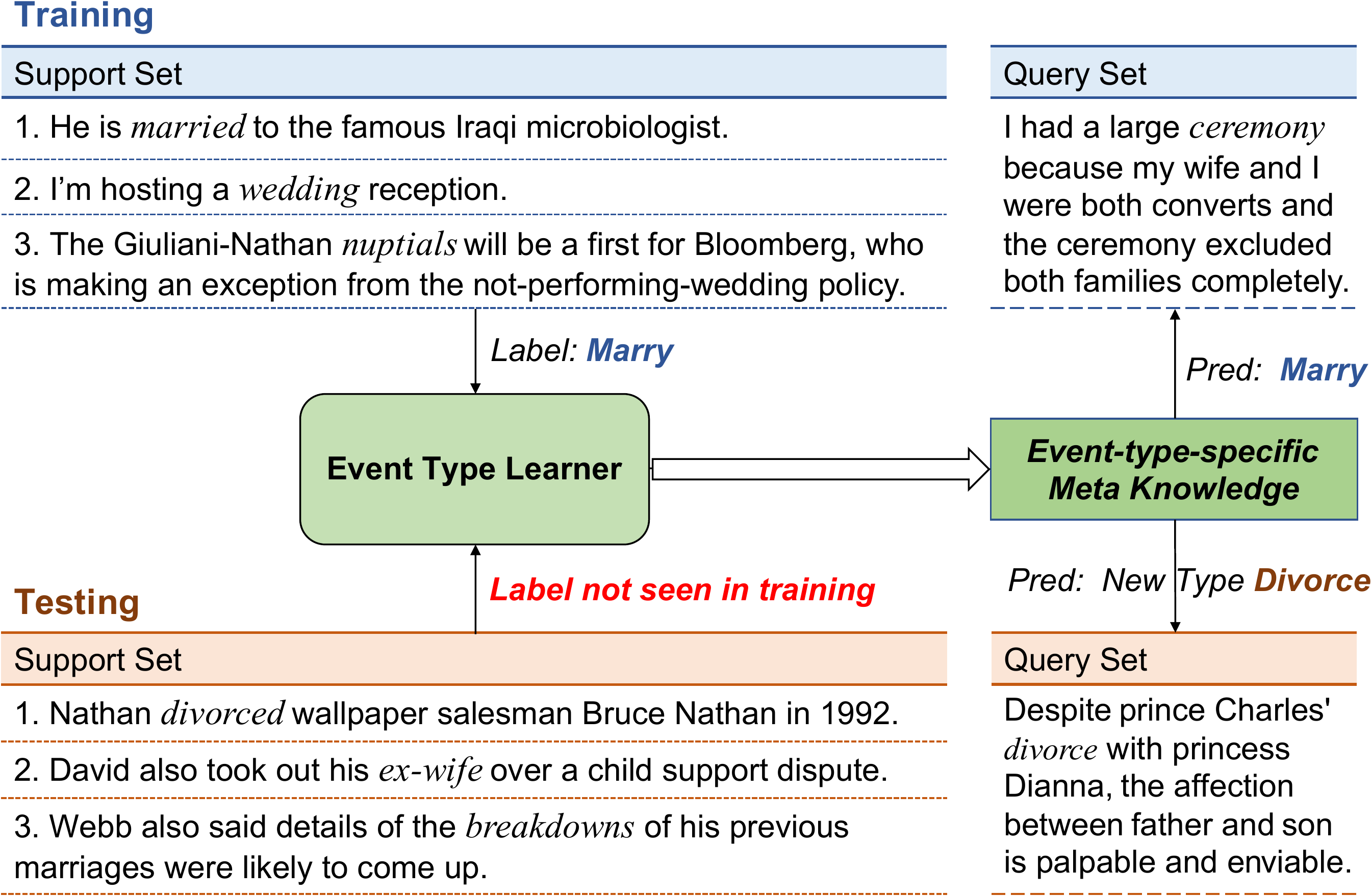}
  \caption{A few-shot (3-shot) event detection example, in which italic words in the support and query set are triggers of events. ‘\emph{Label}' denotes the labeled type of event mentions, and ‘\emph{Pred}' denotes predicted types.
  \label{fig:intro_fsed}}
\end{figure}

In this paper, we revisit the ED task with limited labeled data as an instantiation of FSL problems, \emph{i.e.}, \textbf{Few-Shot Event Detection (FSED)}.
Figure~\ref{fig:intro_fsed} illustrates a few-shot learning example for FSED tasks. Intuitively, the FSED model is analogous to an \emph{Event Type Learner}, which attempts to learn event-type-specific meta knowledge from only few instances in the support set, and apply what it learns to predict the event type of instances in the query set. In a typical meta-learning setting, the \emph{Event Type Learner} is firstly trained in a meta-training step to learn meta knowledge from event types such as \emph{Marry}, afterwards the model is quickly adapted, with only three samples again, to predict the results for new event types such as \emph{Divorce}, which is even not seen in training.

This paper proposes to tackle the problem of FSED in few-shot and meta learning settings.
Non-parametric approaches such as siamese networks \cite{koch2015siamese}, matching networks \cite{vinyals2016matching}, and prototypical networks \cite{snell2017prototypical} are among the most popular models for FSL tasks, due to the properties of simpleness, entirely feed-forward and easy to be optimized. Unlike typical deep learning architecture, non-parametric approaches do not classify instances directly, and instead \emph{learn to compare} in a metric space. For example, prototypical network simply learns a \emph{distance function} to compute a mixture of prototypes for classes. Afterward the encoder compares the new sample with prototypes, and classifies it to the class with the closest prototype \cite{snell2017prototypical}.

Previous studies \cite{snell2017prototypical, gao2019hybrid} demonstrate that selection of distance functions will significantly affect the capacity of prototypical networks, so that the model performance is vulnerable to instance representations. However, due to the paucity of instances in FSL, key information may be lost in noise brought by the diversity of event mentions. Moreover, it is deficient to learn robust contextual representation due to data shortage, particularly for tasks like ED in which learning context-aware embeddings for words and sentences are vital \cite{liu2017exploiting,Liu2018Exploiting}. As a result, in the case of FSED, there is a urgent demand for a more robust architecture that can learn contextual representations for event prototypes from limited instances.

In this work, we propose a Dynamic-Memory-Based Prototypical Network (\textbf{DMB-PN}), which exploits Dynamic Memory Network (DMN) \cite{xiong2016dynamic,kumar2016ask} to learn better prototypes for event types. Differing from vanilla prototypical networks simply computing event prototypes by averaging, which only consume event mention encodings once, DMB-PN, equipped with a DMN, distills contextual information from event mentions for multiple times. Experiments demonstrate that DMB-PN not only deals with sample scarcity better than vanilla prototypical networks, but also performs more robustly when the shot number decreases, referring to the section \label{sec:kshot} on \textbf{K-Shot Evaluations}, and the way number of event types increases, referring to the section \label{sec:nway} on \textbf{N-Way Evaluations}.

Additionally, Dynamic Memory Network is also used to learn event prototypes and sentence encodings in our model. Specifically, we propose to use trigger words as the \emph{questions} in a typical DMN modules to produce the memory vectors, thereby producing  sentence encoding more sensitive to trigger words. As DMN is more advantageous to fully exploit the event instances due to its multi-hop mechanism, DMN-based models are more robust in sentence encodings particularly in few-shot settings as supported by experimental results.


In summary, the main contributions of our work are as follows:
\begin{itemize}
	\item Firstly, we formally define and formulate the new problem of \emph{Few Shot Event Detection}, and produce a new dataset called FewEvent tailored particularly for the problem.

	\item We then propose a new framework called  \emph{Dynamic-Memory-Based Prototypical Network}, which exploits \emph{Dynamic Memory Network} to not only learn better prototypes for event types, but also produce more robust sentence encodings for event mentions.

	\item The experiments show that  prototypical network integrated with memory mechanism outperforms a series of baseline models, particularly when the variety of event types is relatively large and the instance quantity is extremely small, owning to its capability of distilling contextual information from event instances for multiple times.
\end{itemize}

The next section review related work on label-data shortage in event detection and meta-learning in few-shot NLP tasks. Section \ref{sec:method} present the details of DMB-PN architecture. Section \ref{sec:exp} introduce the experiments and evaluation results. Section \ref{sec:con_fuw} make a conclusion of the paper and discusses the future work.

%% file: related_work.tex
\section{Related Work} \label{sec:relw}
\subsection{Sample Shortage Problems in ED Tasks.}
Traditional approaches to the task of EE primarily rely on elaborately-designed features and complicated natural language processing (NLP) tools \cite{mcclosky2011event,li2013joint,hong2011using}. 
Recently, neural-network-based models have shown good performance on EE tasks \cite{nguyen2016modeling,hong2018self,chen2019exploiting,nguyen2018graph,liu2019exploiting}, since \cite{chen2015event} proposed dynamic multi-pooling convolutional neural network (DMCNN) to automatically extract and reserve lexical-level and sentence-level features.
However, these methods rely on large-scale labeled event datasets. 
Considering actual situations, there have been some researches focusing on the shortage of labeled data. 
\cite{nguyen2016two} proposes a CNN-2-STAGE model which uses a two-stage training method to detect event types not seen during training, through effectively transfering knowledge from other event types to the target one.
\cite{peng2016event} develops an event detection and co-reference system with minimal supervision, in the form of a few event examples, by viewing ED tasks as semantic similarity problems among event mentions or among event mentions and ontologies of event types.
\cite{huang2018zero} takes a fresh look at EE by mapping event instances to the corresponding event ontology which holds event structures for each event type.
Besides, there are also some works address the problem of insufficient training data by importing external knowledge\footnote{Note that we do not consider data augmentation by import external pre-train knowledge in this paper and only focus on the few-shot models.}.
\cite{baldini-soares-etal-2019-matching} describes a novel training setup called matching the blanks, and couple it with BERT \cite{devlin-etal-2019-bert} to produce useful relation representations, particularly effective in low-resource regimes.  
\cite{yang-etal-2019-exploring-pre} proposes a method to automatically generate labeled data by editing prototypes and screen out generated samples by ranking the quality.

\subsection{Meta-Learning in Few-Shot NLP Tasks}
Actually, researches about adopting FSL for NLP tasks are extremely limited, and mostly based on metric-based methods.
\cite{gao2019hybrid} formalizes relation classification as a FSL problem, and propose hybrid attention-based prototypical networks for the task.
\cite{yu2018diverse} proposes an adaptive metric learning approach that automatically determines the best weighted combination from a set of metrics for few-shot text classification.
In this paper, we also utilize a metric-based method, prototypical network, to tackle the few-shot event detection tasks.
Besides, model-based methods are also designed for meta-learning to rapidly incorporate new information and remember them. 
Few-shot ED tasks with sparse labeled instances make it vital to make full use of available data, especially contextual information which has been shown effective on ED tasks \cite{liu2017exploiting,nguyen2018graph}.
However, existing methods which utilize context only process the context once. 
\cite{xiong2016dynamic,kumar2016ask} introduce the dynamic memory network (DMN),  exhibiting certain reasoning capabilities in NLP tasks, such as QA, with the multi-hop mechanism.
Inspired by this, \cite{Liu2018Exploiting} proposes the trigger detection dynamic memory network (TD-DMN) to tackle the ED problem by fully exploiting the context in documents.



%% file: method.tex
\begin{figure*}[!hpt]
  \centering
  \includegraphics[scale=0.49]{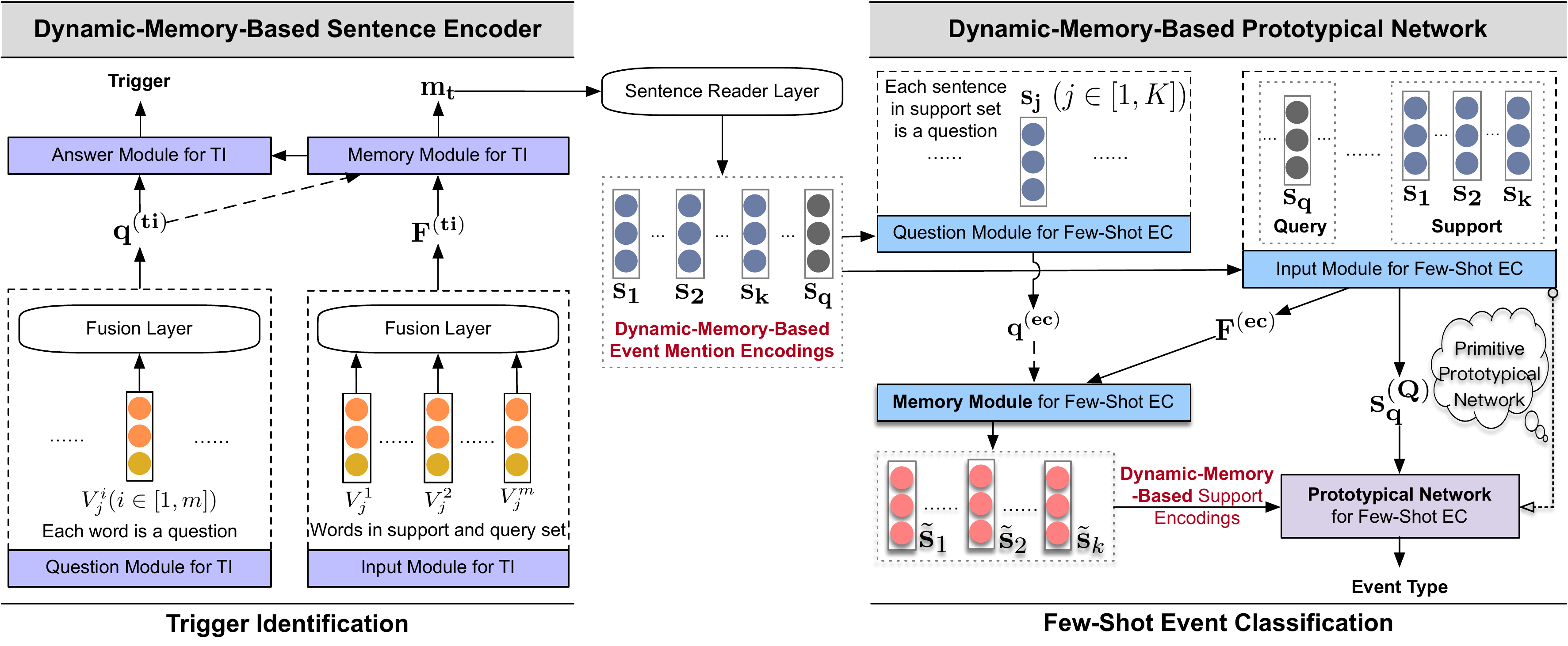}
  \caption{Overview of DMB-PN model, where TI and EC is an abbreviation of trigger identification and event classification respectively. 
  The question in TI is implicitly viewed as ‘‘\emph{Whether the word is a trigger or not?}'', and that in few-shot EC is implicitly viewed as ‘‘\emph{How does this event mention contribute to event prototype learning?}''. 
  Note that the primitive prototypical network directly converts the input module results to prototypical network for few-shot EC, while DMB-PN generates dynamic-memory-based support and query encodings first. 
  \label{fig:model_overview} }
\end{figure*}

\section{Method} \label{sec:method}
This section introduces the general architecture and principal modules of the proposed model. 

\subsection{Problem Formulation}

In this paper, the Few Shot Event Detection (FSED) problem is formulated with typical $N$-way-$K$-shot descriptions. Specifically, our model is given a tiny labeled training set called \emph{support set} $\mathcal{S}$, which has $N$ event types. Each event type has only $K$ labeled samples and $K$ is typically small, for example $1$-shot or $5$-shot. In addition to the \emph{support set}, there is another set called \emph{query set} $\mathcal{Q}$, in which the samples are unlabeled and subject to prediction based only on the observation of few-shot samples in $\mathcal{S}$.


Formally, given an event type set $\mathcal{E}$, the support set $\mathcal{S}$, the query set $\mathcal{Q}$ and few shot task $\mathcal{T}$ are defined as follows.
\begin{equation}
\begin{aligned}
\mathcal{S} & = \{ 
(x^s_1, g^s_1, y^s_1), \cdots, (x^s_i, g^s_i, y^s_i), \cdots \} 
, ~ i \in [1, N \times K]
\\
\mathcal{Q} & = \{
(x^q_1, g^q_1, y^q_1), \cdots, (x^q_i, g^q_i, y^q_i), \cdots \}
, ~ i \in [1, Q] 
\\
\mathcal{T} & = \{\mathcal{S}, \mathcal{Q}\}
\end{aligned}
\label{eq:s_q}
\end{equation}
where $(x^s_i, g^s_i, y^s_i)$ denotes that an event mention instance $x^s_i$ in support set with trigger $g^s_i$ and event type $y^s_i$. Analogously, $(x^q_i, g^q_i, y^q_i)$ denotes an event mention instance in query set, and $Q$ is instance number of query set. 
Each instance $x_j$ is denoted as a word sequence $\{w_j^i | i \in [1, m]\}$, and $m$ is the maximum length of event mentions.

Thus, the goal of few-shot event detection is to gain the capability to predict the type $y$ of a new event in the query set with only observing a small number of events for each $y$ in the support set. Its training process is based on a set of tasks $\mathcal{T}_{train} = \{\mathcal{T}_{i}\}_{i=1}^{Q_{train}}$ where each task $\mathcal{T}_{i} = \{\mathcal{S}_i, \mathcal{Q}_i\}$ corresponds to an individual few-shot event detection task with its own support and query set.
Its testing process is conducted on a set of new tasks $\mathcal{T}_{test} = \{\mathcal{T}_{j}\}_{j=1}^{Q_{test}}$ which is similar to $\mathcal{T}_{train}$, other than that $\mathcal{T}_{j} \in \mathcal{T}_{test}$ should be about event types that have never seen in $\mathcal{T}_{train}$.


\subsection{General Architecture}

Generally,we divide few-shot event detection into two sub-tasks: \emph{trigger identification} and \emph{few-shot event classification}. The overview of our model DMB-PN is shown in Figure~\ref{fig:model_overview}.


In \emph{trigger identification}, a \emph{dynamic-memory-based sentence encoder} is designed to learn event mention encodings and identify triggers.
Given an event mention, each word in it is vectorized to dynamic-memory-based word embedding, and then is identified as a trigger or not based on DMN. 
Specifically, for event mention instance $x_j$, each word $w_j^i$ in it is vectorized to $V_j^i$.
Then the trigger $g_j$ is identified and the sentence encoding $s_j$ is obtained via a dynamic-memory-based sentence encoder $F_x$:
\begin{equation}
[\bm{s_j}, g_j] = F_x (V_j^i), ~ i \in [1, m], j \in [1, N \times K + Q]
\label{eq:sent_enc}
\end{equation}

In \emph{few-shot event classification}, a \emph{dynamic-memory-based prototypical network}, denoted as \emph{M\_Proto}, is proposed to classify events through FSL.
Differing from the primitive prototypical network, the dynamic-memory-based one generates encodings of support set and query set under the architecture of dynamic memory network.
The prototypical network is applied to serve as the answer module of DMN, 
where the event type $y$ is predicted by comparison between query instance encoding $\bm{s_q^{(Q)}}$ and each event prototype $\bm{e_i}~(i \in [1, N])$, denoted by
\begin{equation}
\begin{aligned}
y_i = ~ & M\_Proto(\bm{e_i}, \bm{s_q^{(Q)}}) \\
y = ~ & max(y_i), ~ i \in [1, N], ~ y \in \mathcal{E}
\end{aligned}
\label{eq:final}
\end{equation} 
where $y_i$ denotes the probability of the query instance belongs to the $i_{th}$ event type.

\subsection{Trigger Identification}
\textbf{Input Module for TI.} 
The input module of trigger identification contains two layers: the \emph{word encoder layer} and the \emph{input fusion layer}. The word encoder layer encodes each word into a vector independently, while the input fusion layer gives these encoded word vectors a chance to exchange information between each other.

\emph{Word encoder layer.} 
For the $i_{th}$ word $w_{j}^i$ in the $j_{th}$ event mention $x_j$, the encoding includes two components: (1) a real-valued embedding $\bm{w}_{j}^i \in \mathbb{R}^{d_w}$ to express semantic and syntactic meanings of the word, which are pre-trained via GloVe \cite{pennington2014glove}, and (2) position embeddings to embed its relative distances in the sentence, including distances from $w_{j}^i$ to the beginning and ending of the sentence, as well as the sentence length, with three $d_p$-dimensional vectors, and then concatenate them as a unified position embedding $\bm{p}_{j}^i \in \mathbb{R}^{3 \times d_p}$.

We then achieve a final input embedding $\bm{V}_{j}^i$ for each word by concatenating its word embedding and position embedding. 
\begin{equation}
\bm{V}_{j}^i = [\bm{w}_{j}^i, \bm{p}_{j}^i]
\end{equation}

\emph{Input fusion layer.} 
Given $\bm{V}_{j}^i$, we generate fact vectors $\bm{F^{(ti)}}$ with a Bi-GRU.
\begin{equation}
	\bm{F^{(ti)}} = \{\bm{f}_{j}^i | \bm{f}_{j}^i = {BiGRU}_w (\bm{V}_{j}^i), 
	i \in [1, m], j \in [1, N \times K + Q] \}
\label{eq:facts_word}
\end{equation}
where $m$ is the maximum sentence length.

\textbf{Question Module for TI.}
Analogously, the question module encodes the question into a distributed vector representation. 
In the task of trigger identification, each word in the input sentence could be deemed as the question. The question module of trigger identification treats each word in the event mention as implicitly asking a question ‘‘\emph{Whether the word is the trigger or not?}''.
The intuition here is to obtain a vector that represents the question word. 
Given encoding of the $i_{th}$ word in the $j_{th}$ sentence, $\bm{V}_{j}^i$, the question GRU generates hidden state $\bm{q}_{j}^i$ by a Bi-GRU.
The question vector $\bm{q}_j$ for the $j_{th}$ sentence is a combination of all hidden states.
\begin{equation}
\bm{q}_j^{(ti)} = \{ \bm{q}_{j}^i | i \in [1, m] \} 
\end{equation}



\textbf{Answer Module for TI.} 
The answer module predicts the trigger in an event mention from the final memory vector of the \textbf{Memory Module for TI}, which will be introduced in \emph{Memory Module for TI and Few-Shot EC} of Section \ref{sec:fsec}.
We employ another GRU whose initial state is initialized to the last memory $d_0 = m_{\mathcal{T}}$. At each timestep, it takes the question $\bm{q}_j$, last hidden state $d_{t-1}$, as well as the previously predicted output $y_{t-1}^{(ti)}$ as input.
\begin{equation}
	y_t^{(ti)} = \hat{g}_t = softmax (W^{(ti)} d_t)
\end{equation}
\begin{equation}
	d_t = GRU([y_{t-1}^{(ti)}, \bm{q}_j], d_{t-1})
\end{equation}

The output of trigger identification are trained with cross-entropy error classification, and the loss function for trigger identification is denoted by
\begin{equation}
\begin{small}
	L_{TI} = -[y^{(ti)} \log \hat{y}^{(ti)} + (1 - y^{(ti)}) \log (1 - \hat{y}^{(ti)})]
\label{eq:loss_ti}
\end{small}
\end{equation}

\textbf{Sentence Reader Layer.}
The sentence reader layer is responsible for embedding the words into a sentence encoding, where the words are embedded through the memory module of trigger identification.
We obtain scalar attention weight for each word in a sentence by feeding $\bm{m}_{j}^i$ generated by the \textbf{Memory Module for TI} into a two-layer perceptron and going through a softmax.

\begin{equation}
	\alpha_{j}^i = softmax( \tanh(\bm{m}_{j}^i \cdot W_{s_1}) \cdot W_{s_2} )
\end{equation}

Then we denote sentence representation $\bm{s}_j$ by
\begin{equation}
	\bm{s}_j = \sum_{i=0}^{n-1} \alpha_{j}^i \bm{f}_{j}^i
\end{equation}

\subsection{Few-Shot Event Classification}\label{sec:fsec}

\textbf{Input Module for Few-Shot EC.}
The input module of few-shot event classification is after a sentence integration layer and contains an input fusion layer. 
The sentence integration layer integrates sentences into support set and query set respectively.
The input fusion layer gives these sentence encodings in support set a chance to exchange information between each other.

\emph{Sentence integration layer.}
The support set and query set encoding are denoted by 
\begin{equation}
\begin{aligned}
	\bm{s}^{(S)} & = \{ \bm{s}_{ij} | \bm{s}_{ij} = \bm{s}_k \land k \in [1, N \times K], i \in [1, N], j \in [1, K] \} \\
	\bm{s}^{(Q)} & = \{ \bm{s}^{(Q)}_q | \bm{s}^{(Q)}_q = \bm{s}_q \land q \in [1, Q] \}
\end{aligned}
\label{eq:support_query_set_enc}
\end{equation}

\emph{Input fusion layer.}
The input fusion operation for sentences is similar to that for words, shown in Equation~\eqref{eq:facts_word}.
We generate the fact vectors for sentences in support set by a Bi-GRU.
\begin{equation}
\bm{F^{(ec)}} = \{ \bm{f_j} | \bm{f_j} = BiGRU_s(\bm{s}^{(S)}), ~ j \in [1, N \times K] \}
\end{equation}


\textbf{Question Module for Few-Shot EC.}
In the task of few-shot event classification, each event mention could be deemed as the question. The question module of few-shot event classification treats each event mention as implicitly asking a question ‘‘\emph{How does this event mention contribute to event prototype learning?}''

The question vector for event mention is obtained by feeding sentence encodings in support set to a Bi-GRU:
\begin{equation}
	\bm{q^{(ec)}} = \{ \bm{q}_j | \bm{q}_j = {BiGRU}_q (\bm{s}^{(S)}), ~ j \in [1, N \times K]  \} 
\label{eq:ques_sent}
\end{equation}

\textbf{Memory Module for TI and Few-Shot EC.} 
The memory module for trigger identification and few-shot event classification are almost the same, except the inputs are word encodings and event mention encodings respectively. Given a collection of inputs, the episodic memory module chooses which parts of inputs to focus on through attention mechanism. It then produces a new ‘‘memory'' vector considering the question as well as the previous memory. At each iteration, the memory module is able to retrieve new information which were thought to be irrelevant in previous iterations.

\begin{figure}[!hpt]
  \centering
  \includegraphics[scale=0.4]{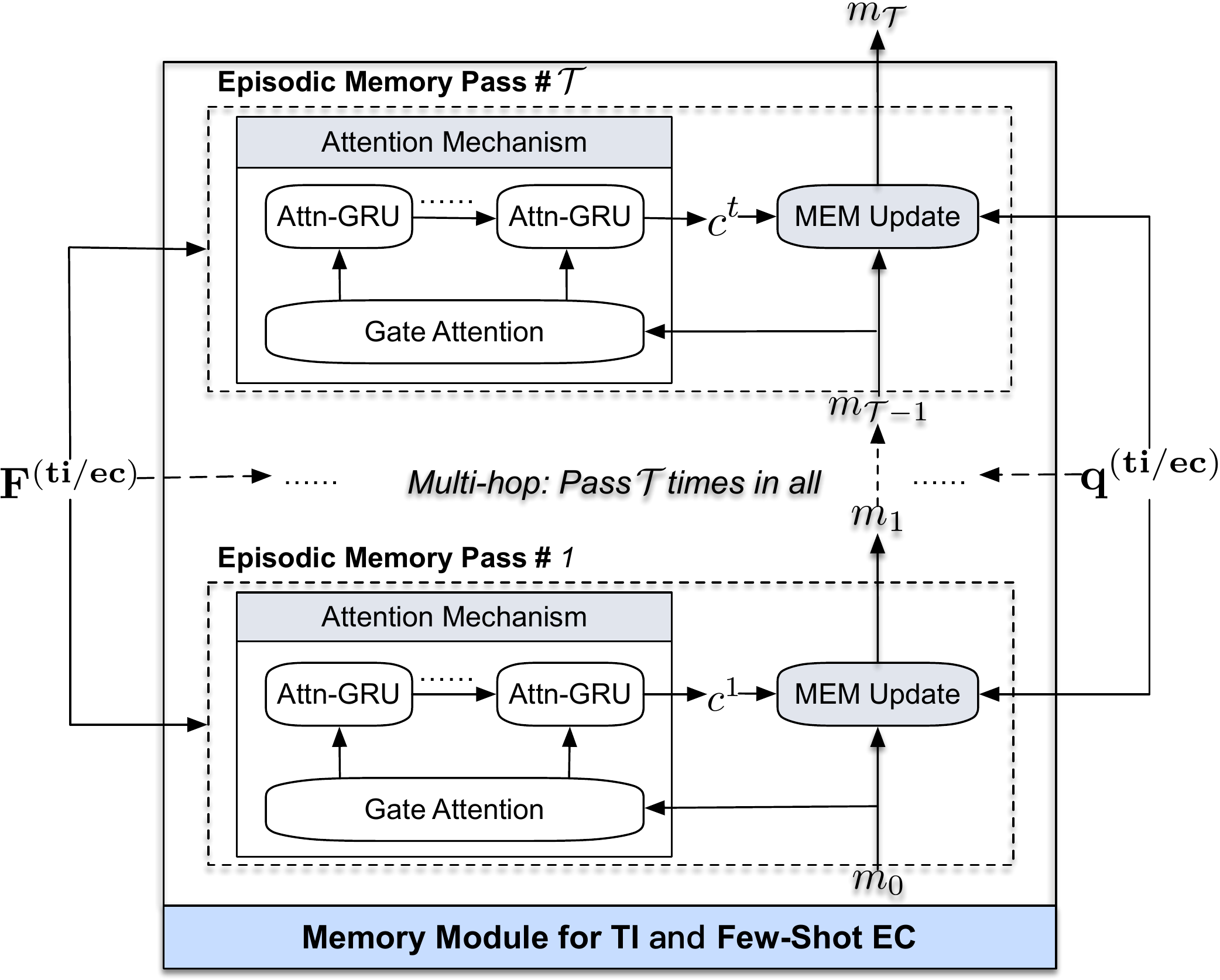}
  \caption{Architecture of DMB-PN memory module. 
  $\bm{F}$ denote facts of input, $q$ denotes question vector, $c$ denotes candidate facts, and $m$ denotes memory. 
  \label{fig:model_MemM}}
\end{figure}

Specifically, the memory module contains three components: the \emph{attention gate}, the \emph{attentional GRU} \cite{xiong2016dynamic}, and the \emph{memory update gate}. We present its structure in Figure~\ref{fig:model_MemM}.

\emph{Attention gate.} The attention gate determines how much the memory module should pay attention to each fact, given the facts $\bm{F} = \{\bm{f_1}, \cdots, \bm{f_n}\}$, the question $\bm{q}$, and the acquired knowledge stored in the memory vector $\bm{m}_{t-1}$ from the previous step. The three inputs are transformed by:

\begin{equation}
\bm{u} = [\bm{F} \circ \bm{q}, |\bm{F} - \bm{q}|, \bm{F} \circ \bm{m}_{t-1}, |\bm{F} - \bm{m}_{t-1}|]
\end{equation}

where ‘‘$,$'', $\circ$, $-$ and $|\cdot|$ are concatenation, element-wise product, subtraction and absolute value respectively. The first two terms measure similarity and difference between facts and the question, and the last two terms comparing facts with the last memory state.

Let $\bm{a}$ of size $n_I$ denotes the generated attention vector. The $i_{th}$ element in $\bm{a}$ is the attention weight for fact $\bm{f}_i$. $\bm{a}$ is obtained by transforming $\bm{u}$ using a two-layer perceptron:
\begin{equation}
	\bm{a} = softmax ( \tanh( \bm{u} \cdot W_{m_1} ) \cdot W_{m_2} )
\label{eq:attn_facts}
\end{equation}
where $W_{m_1}$ and $W_{m_2}$ are parameters of the perceptron.

\emph{Attentional GRU.}
The attentional GRU takes facts $\bm{F}$, fact attention $\bm{a}$ as input and produces context vector $\bm{c}$.
\begin{equation}
	\bm{f}_t = \bm{a} \circ \bm{f}_t
\end{equation}
\begin{equation}
	\bm{h}_t = GRU( \bm{f}_t, \bm{h}_{t-1} )
\end{equation}
Context vector $\bm{c}$ is the final hidden state of attention based GRU:
\begin{equation}
	\bm{c} = \bm{h}_T
\end{equation}

\emph{Memory update gate.} 
The episodic memory for passing $\mathcal{T}$ times is computed by
\begin{equation}
	\bm{m}_{\mathcal{T}} = GRU(\bm{c}, \bm{m}_{\mathcal{T}-1})
\end{equation}
and the new episode memory state is calculated by
\begin{equation}
	\bm{m}_{\mathcal{T}} = relu(W_t[\bm{m}_{\mathcal{T}-1}, \bm{c}, \bm{q}] + b)
\label{eq:memory_of_facts}
\end{equation}
where ‘‘$,$" is concatenation operator, $W_t \in \mathbb{R}^{n_H \times n_H}$, and $b \in \mathbb{R}^{n_H}$.

\textbf{Memory-Based Prototypical Network for Few-Shot EC.}
The main idea of prototypical networks for few-shot event classification is to use a feature vector, also named a prototype, to represent each event type. The traditional approach to compute the prototype is to average all the instance embeddings in the support set to produce the event type. In this paper, we apply memory-based mechanism to produce event prototypes.

In practice, event mentions for an event type can be of great discrepancy, and the huge diversities among instances may result in inaccurate representation of events. In order to obtain more precise event prototype $\bm{e}_k$, we encode each event mention $\tilde{\bm{s}}_k ~ (k \in [1, N \times K])$ in support set by making interaction with other event mentions of the same event type, which is calculated by Equation~\eqref{eq:memory_of_facts}. 

We then compute probabilities of event types for the query instance $\bm{s}^{(Q)}_q$ (Equation~\eqref{eq:support_query_set_enc}) as follows
\begin{equation}
	P(y = y_k) = \frac{exp(-|| \bm{s}^{(Q)}_q - \bm{e}_k ||)}{\sum_{j=1}^N exp(-|| \bm{s}^{(Q)}_q - \bm{e}_j ||)}
\end{equation}
where $||\cdot||$ denotes Euclidean	distance.

\emph{Loss function.} 
We adopt the cross entropy function as the cost function for few-shot event classification, calculated by
\begin{equation}
	L_{EC} = - \sum_{k=1}^{N} y^{(ec)} \log P(y = e_k)
\end{equation}

The final loss function for few-shot event detection is a weighted sum of trigger identification loss and few-shot event classification loss, denoted by
\begin{equation}
	L = \lambda L_{TI} + (1 - \lambda)L_{EC} \label{eq:loss}
\end{equation}
where $\lambda$ is a hyper-parameter, and we set it to $0.5$ in this paper.

%% file: experiment.tex
\section{Experiments} \label{sec:exp}
The experiments seek: (1) to compare the dynamic-memory-based prototypical network with a series of combinations of sentence encoding models and metric models; (2) to assess the effectiveness of memory-based models from the perspective of $K$-shot evaluations and $N$-way evaluations respectively in different $N$-way-$K$-shot settings; (3) to provide the evidence for that dynamic-memory-based approaches are more feasible to learn contextual representations for both event prototypes and event mentions from limited instances.

\input{tab/T_Exp_EC_Accuracy_Model}

\subsection{Datasets}
FSED task should be trained and tested on few-shot event detection datasets  as few-shot tasks in other research areas, while there are not existing FSED datasets. Thus we evaluate our models on a newly-generated dataset tailored particularly for few-shot event detection called \emph{FewEvent}. In general, it contains $70,852$ instances for $19$ event types graded into $100$ event subtypes in total, in which each event type is annotated with about $700$ instances on average. \emph{FewEvent} was built in two different methods:

\begin{itemize}
  \item We first scale up the number of event types in existing datasets, including the ACE-2005 corpus\footnote{http://projects.ldc.upenn.edu/ace/}, and TAC-KBP-2017 Event Track Data\footnote{https://tac.nist.gov/2017/KBP/Event/index.html}.

  \item We then import and extend some new event types based on an automatically-labeled event data\footnote{https://github.com/acl2017submission/event-data} \cite{chen2017automatically}, from Freebase \cite{Bollacker2008Freebase} and Wikipedia \cite{Witten2008Learning}, constrained to specific domains such as music, film, sports, education, etc.
\end{itemize}

The \emph{FewEvent} dataset is now released and published at \url{https://github.com/231sm/Low_Resource_KBP}, including details of event types and their instance quantity.

In our experiment settings, $80$ event types are selected for training, $10$ for validation, and the rest $10$ event types for testing. Note that there are no overlapping types between training and testing sets, following the typical few-shot settings.


\subsection{Baselines and Settings}
Comparisons are performed against two types of baselines including sentence encoding models and metric learning models. For sentence encoder baselines, we consider 
four models including CNN \cite{kim2014convolutional,zeng2014relation}, Bi-LSTM \cite{huang2015bidirectional}, Self-Attention model \cite{NIPS2017_7181}, and DMN \cite{kumar2016ask,xiong2016dynamic}. For metric-learning baselines, we mainly consider two commonly-used metric models, \emph{i.e.}, Matching Networks \cite{vinyals2016matching} and Prototypical Networks \cite{snell2017prototypical}, as well as our proposed Memory-based Prototypical Network. Combining these two sets of models in pairs, we obtain $4 \times 3 = 12$ baselines. The combination of DMN and Memory-based Prototypical Network is our proposed model, denoted as DMB-PN.

With regard to settings of training process, stochastic gradient descent (SGD) \cite{Ketkar2014Stochastic} optimizer is used, with $30,000$ iterations of training and $2,000$ iterations of testing.
The dimension of memory units, word embedding and position embedding are set to $50$, $50$ and $30$ respectively. The number of memory module passing is $3$. In DMB-PN, a dropout rate of $0.2$ is used to avoid over-fitting, and the learning rate is $1 \times 10^{-3}$. We evaluate the performance of event detection with $Accuracy$ and $F1~Score$.
%

\subsection{General Comparisons} 
As shown in Table~\ref{tab:exp_effect_dmb_model}, we compare \emph{accuracies}, \emph{F1 scores} and \emph{accuracy margins} among different combinations of sentence encoders and metric models. 

A general inspection reveals that Prototypical Network (PN) outperforms Matching Network (MN) when sentence encoders are the same in almost all N-way-K-shot settings. A possible explanation for this might be that PN learns to compare between a query instance and an event prototype, i.e.,\emph{instance-to-prototype matching}, whereas MN compares between instances in the support set and query set,i.e.,\emph{instance-to-instance matching}. Instance-to-instance matching is more susceptible to noises in metric-computation than instance-to-prototype comparing. If there are many outlier instances in the support set, instance-to-instance matching will introduce more noises. This result confirms previous study that prototype learning reduces noises introduced by instance randomness \cite{snell2017prototypical}. 

Notably, the best result is achieved by DMB-PN, a prototypical network incorporated with a DMN. This result echos the statement that \emph{dynamic-memory-based prototypical network learns better prototypes than simply averaging over the instances of the support sets, owning to its capability of distilling contextual information from event instances for multiple times and in an incremental way.}

\subsection{K-Shot Evaluations} \label{sec:kshot}
This section is primarily intended to assess the effectiveness of \emph{memory-based models} from the perspective of $\emph{K}$-$\emph{shot}$ in different $N$-way-$K$-shot settings with the same-way-number settings, such as $5$-way-$5$-shot, $5$-way-$10$-shot and $5$-way-$15$-shot. As shown in Table~\ref{tab:exp_effect_dmb_model}, the effect is reflected by the variance of accuracy margin, denoted as \emph{(+m)} in the brackets and defined as the margin between the accuracy of the worst baseline model and that of the current model under inspection. We report the analysis for both metric models and sentence encoders respectively.

\subsubsection{On Dynamic-Memory-Based Prototypical Networks} When the sentence encoders are the same, we could observe that DMB-PN achieves the best accuracy margin in both $5$-way and $10$-way settings. Further inspection reveals that the accuracy margin of DMB-PN increases as the shot number decreases, indicating that the model performs even better when the shot number is relatively small. In contrast, for other metric-based models such as prototypical network, the margin does not always increases steadily when shot number decreases, \emph{e.g.}, the margin for CNN-PN with a combination of \emph{CNN} and \emph{Proto} increase first then decrease. The possible reason for this is that memory mechanism in DMB-PN has less dependence on the quantity of instances. It is still capable of learning distinguishable event prototypes even the number of instances in the support set is very small. These results corroborate the statement that \emph{the prototypical network integrated with memory mechanism is more applicable to the few-shot classification tasks, particularly when the instance quantity is relatively small.}

\subsubsection{On Dynamic-Memory-Based Sentence Encoders} Given the same metric models such as \emph{M-Proto}, the DMN-based encoders achieves the best margin, \emph{e.g.}, $+15.38$ for DMB-PN. A further inspection reveals that, all models with DMN-based encoder such as \emph{DMN-PN} or \emph{DMN-MN}, the margin \emph{increases} as shots number \emph{decreases}, whereas models with other encoders are different. These results indicate that DMN-based models are more robust in learning sentence encodings, particularly when the shot number is relatively small. A possible explanation for this might be that the multi-hop mechanism of memory-based models is more advantageous to fully utilize the limited instances, while the other sentence encoders only consume the training samples for once.

\begin{figure}[!htbp]
  \vspace{-2mm}
  \centering
  \includegraphics[scale=0.125]{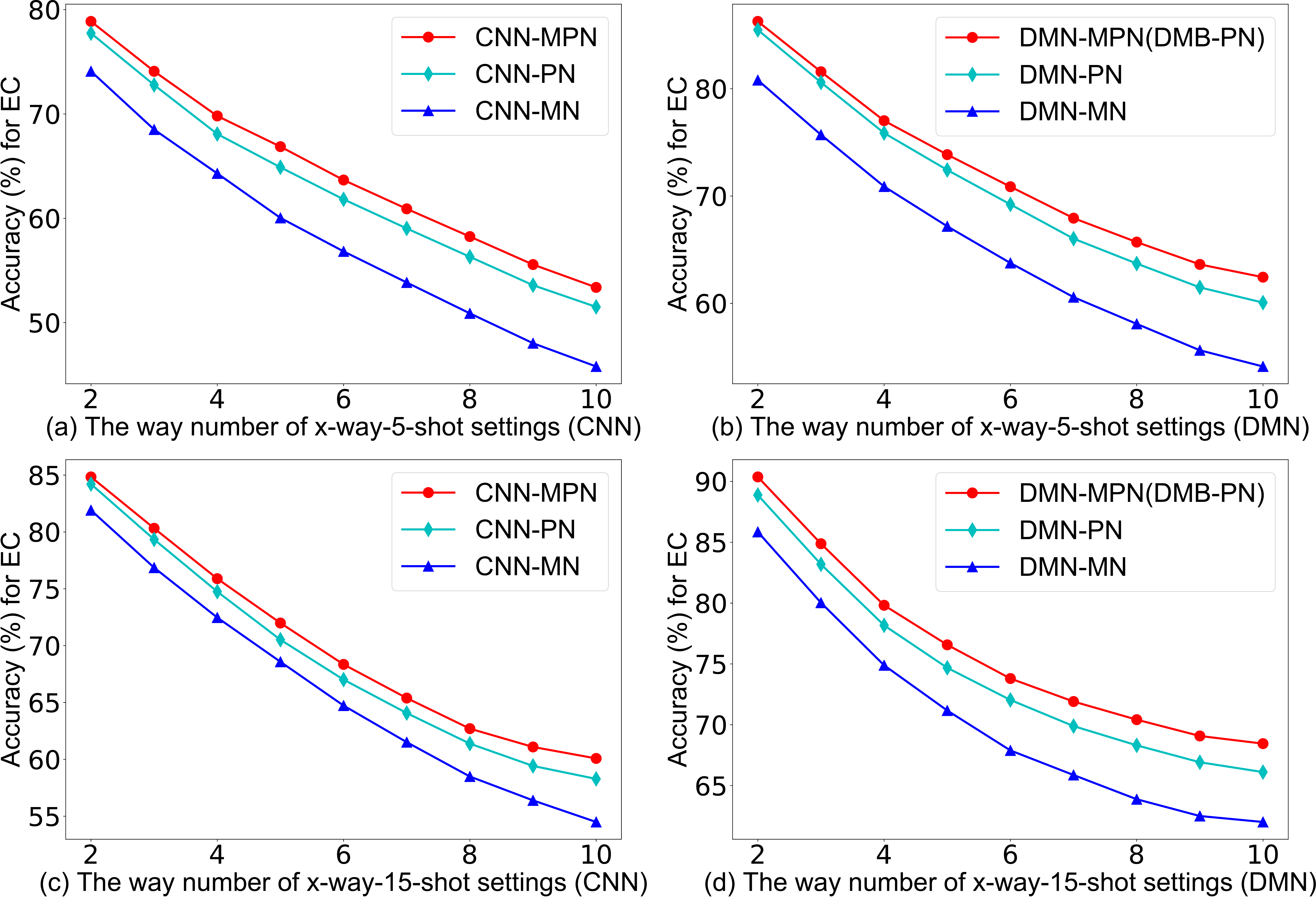}
  \vspace{-7mm}
  \caption{N-way evaluations with fixed shot numbers. (a) and (b) illustrate the variation tendency of accuracy in the $5$-shot setting for models with CNN encoders and DMN encoder respectively. (c) and (d) illustrate the results for $15$-shot setting while the way number increases. \label{fig:exp_comp_metric}}
  \vspace{-6mm}
\end{figure}

\subsection{N-Way Evaluations} \label{sec:nway}
This section is mainly intended to assess the effectiveness of \emph{memory-based models} from the perspective of $\emph{N}$-$\emph{way}$ in $N$-way-$K$-shot settings with the same shot numbers, as illustrated in Figure~\ref{fig:exp_comp_metric}. Generally, the accuracy decreases as the way number increases when the shot number is fixed, which is in accordance with the expectation as larger number of ways results in wider variety of event types to be predicted, which increases the difficulty of correct classifications. We can further observe that memory-based models such as \emph{CNN-MPN} performs better than vanilla prototypical networks which further overtakes matching networks, and the margins among them increases as way number increases. These results indicate that the memory-based prototypical network is more robust to the number of ways as multi-hop mechanism in memory networks contributes to the learning of more distinguishable event prototypes.

\begin{figure}[!htp]
  \vspace{-3mm}
  \centering
  \includegraphics[scale=0.2]{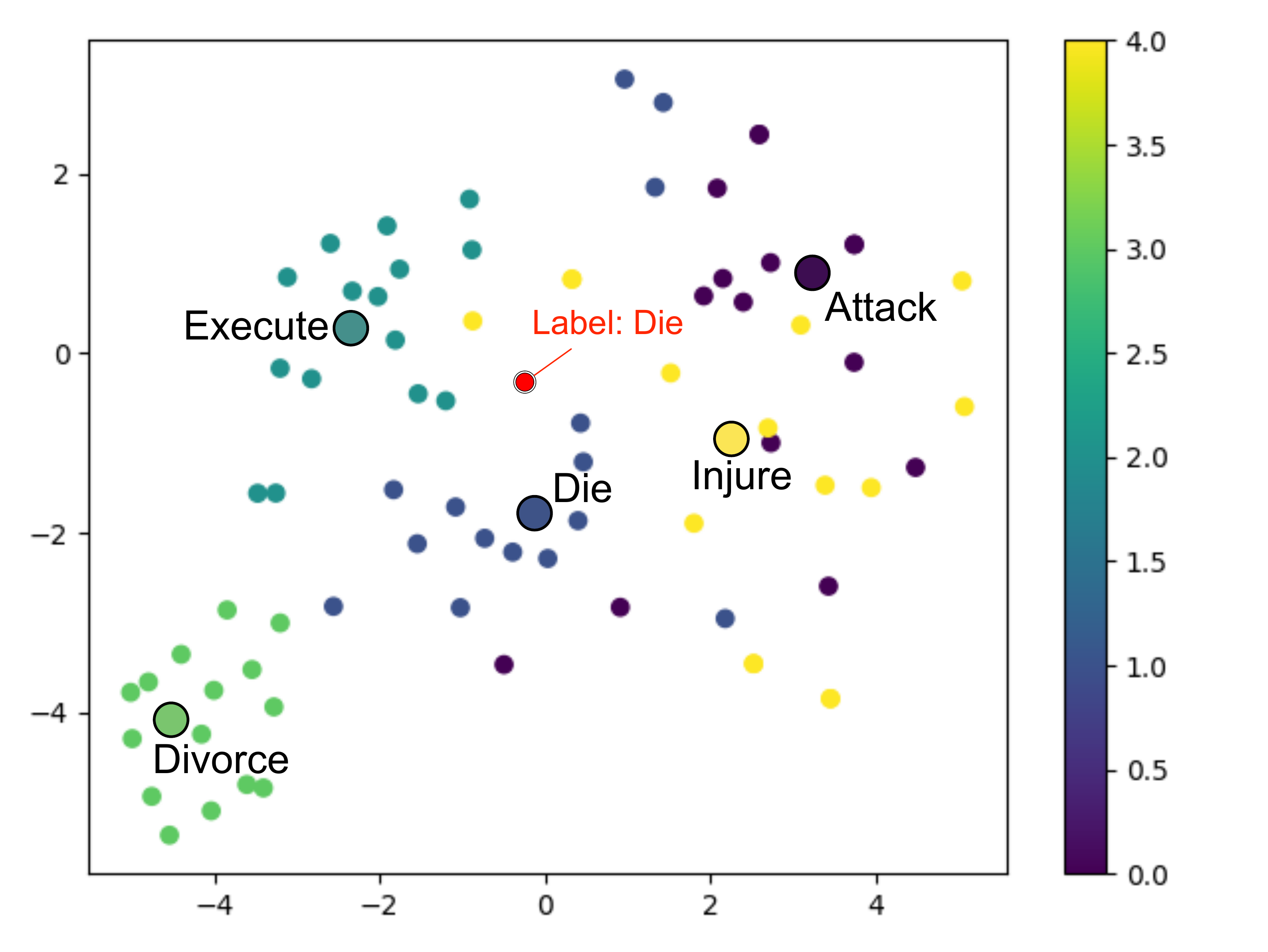}
  \vspace{-5mm}
  \caption{Visualization of event prototypes, support set, and query instance of DMB-PN, in $5$-way-$15$-shot event detection tasks. Note that the five bigger dot with outlines denote event prototypes, and events of the same type are marked in the same color. The red dot represents a query instance. 
  \label{fig:exp_sents_emb} }
  \vspace{-5mm}
\end{figure}

\begin{figure*}[!htp]
  \centering
  \includegraphics[scale=0.36]{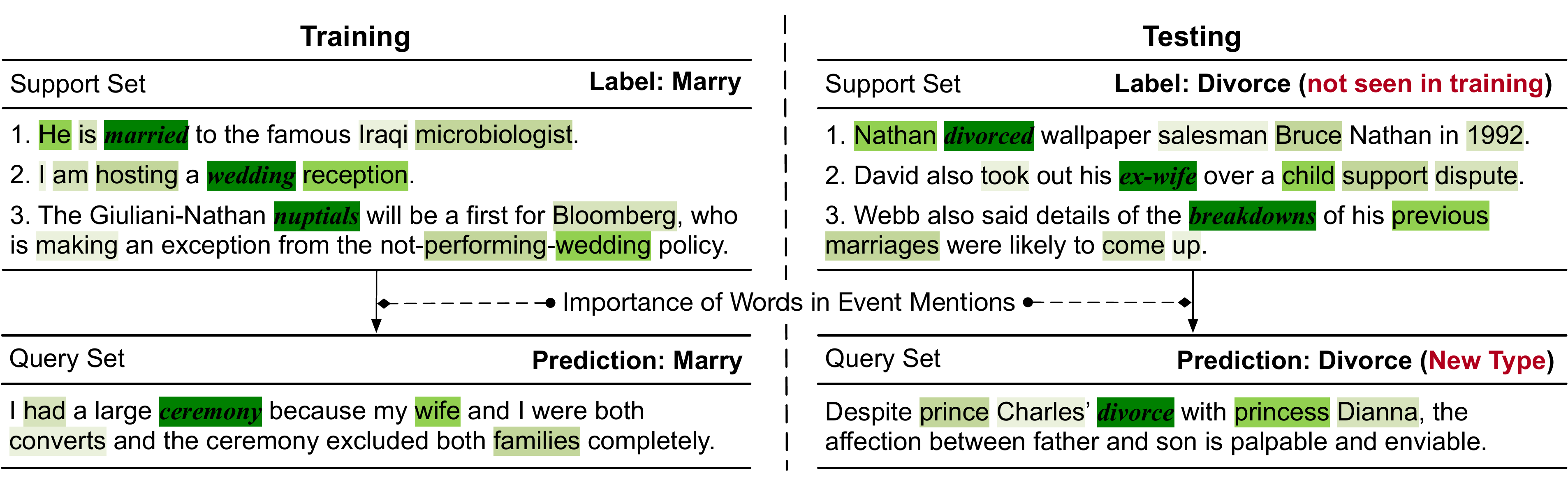}
  \vspace{-5mm}
  \caption{Visualization of word attentions in event mention encodings, where instances of support and query set in training and testing are shown respectively. Note that the attention value becomes smaller as green becomes lighter, and triggers are marked in bold Italic. We only colorize words with top $5$ large attention values of event mention encodings for conciseness. \label{fig:exp_word_attns}}
  \vspace{-3mm}
\end{figure*}

\subsection{Case Study}
This section reports case studies on the learning results of event prototypes and the effectiveness of the model to learn meta knowledge from sentences of event instances.
\subsubsection{On Event Prototypes}
Figure~\ref{fig:exp_sents_emb} visualizes several samples of event prototypes and event mention encodings generated by DMB-PN. One interesting finding is that it is obvious that the distance between the query instance and the prototypes are fairly distinguishable, whereas, it is hardly to distinguish the distances between the query instances \emph{Die}, marked in red, and the surrounding instances of \emph{Execute}, \emph{Injure} and \emph{Die}, marked in dark green, yellow and purple respectively, as the distance distributions are very similar. We can easily predict that it belongs to \emph{Die} as it is closer to the event prototype of \emph{Die}. This example supports the statement that \emph{DMB-PN has advantages in few-shot event detection tasks, especially with instances close in the vector space, through generating distinguishable event prototypes.}



\vspace{-1.5mm}
\subsubsection{On Trigger Detection}
To assess the effectiveness of learning and converting event-type-specific meta knowledge from sentence instances, we visualize word attentions obtained from event mention encoding via DMB-PN, as shown in Figure~\ref{fig:exp_word_attns}.
Apparently, in the process of training, triggers in each event mention tends to achieve higher attention value than other words, and similar results are also obtained during testing, indicating that DMB-PN can effectively detect triggers in event mentions.
Additionally, a further inspection into examples in training reveals that other high-lightened words are participants involved in an event, or provide important clues to ED task, which can be seen as \emph{arguments} \cite{chen2015event}. 
In the process of testing, the \emph{arguments} of each event mention also achieve higher attention. 
For example, in the event mention of ‘‘Nathan divorced wallpaper salesman Bruce Nathan in 1992.'' whose trigger is ‘‘divorced'', DMB-PN considers ‘‘divorced'', ‘‘Nathan'', ‘‘Bruce'', ‘‘1992'' and ‘‘salesman'' as the top $5$ words to be valued, among which the latter four words all describe the event of \emph{Divorce}. This observational study suggests that DMB-PN is capable of capturing both \emph{trigger} and \emph{arguments} information, thereby generating more accurate sentence encoding and capturing more valuable information from limited labeled training data.
It can therefore be assumed that DMB-PN is able to assimilate more valuable meta knowledge from the few-shot samples, and transfer more event-type-specific knowledge for few-shot event classifications.

\vspace{-2mm}
\subsection{Parameter Analysis}
In this section, we intend to study the effect of loss ratio, $\lambda$ in Equation~\eqref{eq:loss}, on trigger identification.

\begin{figure}[!htbp]
  \vspace{-3mm}
  \centering
  \includegraphics[scale=0.12]{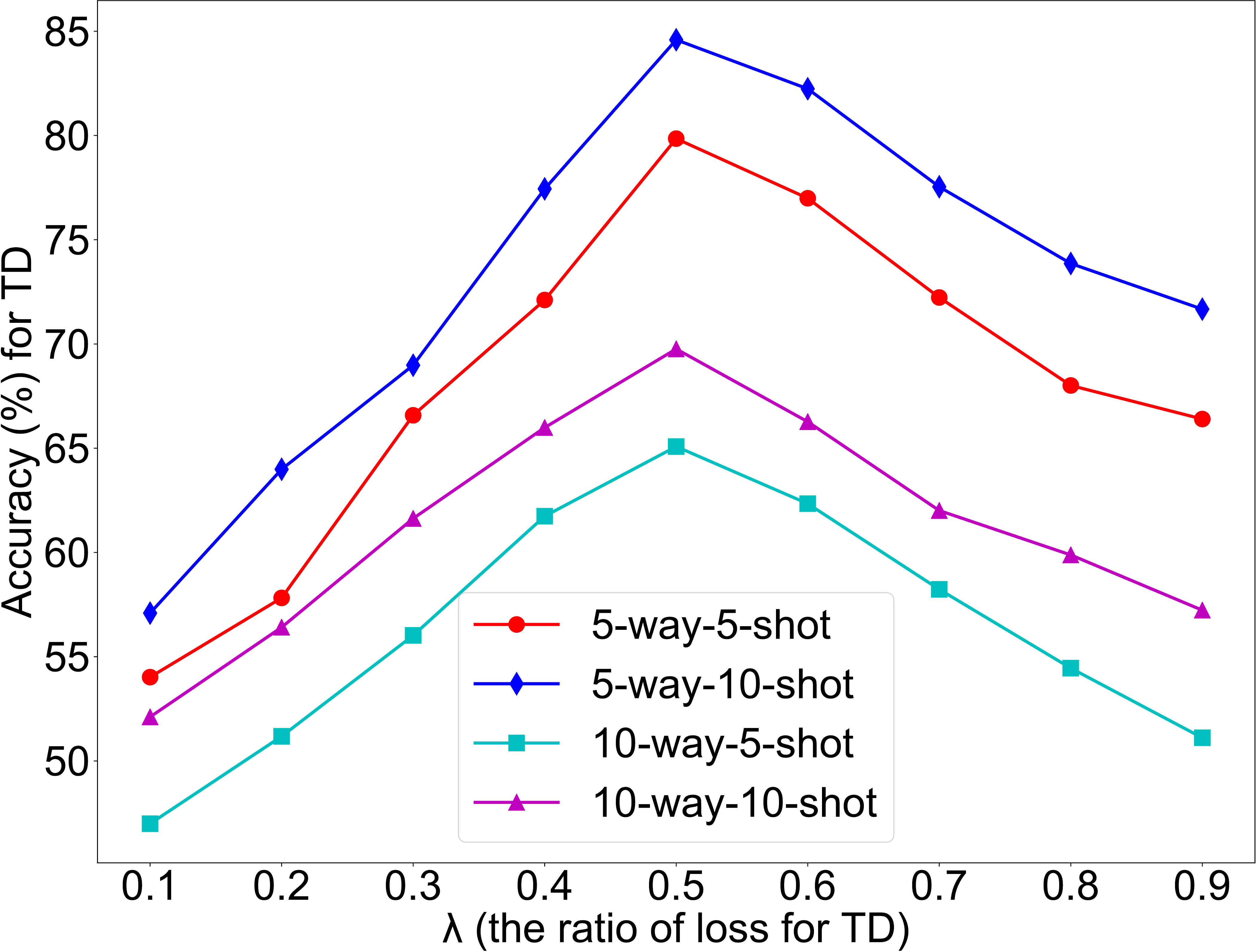}
  \vspace{-4mm}
  \caption{The trigger identification accuracy of DMB-PN model in different few-shot tasks. \label{fig:exp_td}}
  \vspace{-4mm}
\end{figure}

Seen in Figure~\ref{fig:exp_td}, as $\lambda$ increases, the performance of trigger identification increases first and then decreases. When $\lambda$ reaches to $0.5$, the best performance is achieved. This is also why we choose $0.5$ as the value of hyperparameter.
Besides, in general, when $\lambda$ is larger, the performance of trigger identification is better than when it is small. Intuitively, the bigger $\lambda$ implies the model more likely to learn more precise trigger identification results, but not always.
In DMB-PN, the training of trigger identification and few-shot event classification interact with each other, and the final results are actually combination of them. Therefore, we select median as the loss ratio of trigger indentification, and results in Figure~\ref{fig:exp_td} also demonstrate its effectiveness.
Therefore, DMB-PN has its advantage naturally to integrate trigger identification and few-shot event classification, making them influence mutually while both achieve good performance.

%% file: tab/T_Exp_EC_Accuracy_Model.tex
\begin{table*}[!htbp]
\centering
\small
\resizebox{\linewidth}{!}{
	\begin{tabular}{ c | c | c ||  c  c  c || c  c  c}
	\toprule

	\multirow{2}*{Model} & \multirow{2}*{Encoder} & \multirow{2}*{Metric} & 5-Way-5-Shot & 5-Way-10-Shot & 5-Way-15-Shot & 10-Way-5-Shot & 10-Way-10-Shot & 10-Way-15-Shot \\
	\cline{4-9}
	& & & 
	$F1~|~Acc ~(+m)$ & $F1~|~Acc ~(+m)$ & $F1~|~Acc ~(+m)$ & $F1~|~Acc ~(+m)$ & $F1~|~Acc ~(+m)$ & $F1~|~Acc ~(+m)$ \\
	\midrule

	BRN-MN & Bi-LSTM & Match & 
	$58.19~|~58.48 ~(+0.00)$ & $61.26~|~61.45 ~(+0.00)$ & $65.55~|~66.04 ~(+0.00)$ & 
	$46.43~|~47.62 ~(+1.82)$ & $51.97~|~52.60 ~(+1.93)$ & $56.27~|~56.47 ~(+1.98)$ \\

	CNN-MN & CNN & Match & 
	$59.30~|~60.04 ~(+1.56)$ & $64.81~|~65.15 ~(+3.70)$ & $68.35~|~68.58 ~(+2.54)$ &
	$44.85~|~45.80 ~(+0.00)$ & $50.14~|~50.67 ~(+0.00)$ & $54.13~|~54.49 ~(+0.00)$ \\

	SAT-MN & Self-Attn & Match & 
	$63.05~|~64.32 ~(+5.84)$ & $66.93~|~67.62 ~(+6.17)$ & $69.13~|~69.80 ~(+3.76)$ &
	$49.95~|~51.17 ~(+5.37)$ & $55.62~|~56.68 ~(+6.01)$ & $60.18~|~60.53 ~(+6.04)$ \\

	DMN-MN & DMN & Match & 
	$66.09~|~67.18 ~(+\underline{8.70})$ & $68.92~|~69.33 ~(+\underline{7.88})$ & $70.88~|~71.17 ~(+\underline{5.13})$ &
	$52.81~|~54.12 ~(+\underline{8.32})$ & $58.04~|~58.38 ~(+\underline{7.71})$ & $61.63~|~62.01 ~(+\underline{7.52})$ \\

	\midrule

	BRN-PN & Bi-LSTM & Proto &
	$62.42~|~62.72 ~(+4.24)$ & $64.65~|~64.71 ~(+3.26)$ & $68.23~|~68.39 ~(+2.35)$ &
	$53.15~|~53.59 ~(+7.79)$ & $55.87~|~56.19 ~(+5.52)$ & $60.34~|~60.87 ~(+6.38)$ \\ 

	CNN-PN & CNN & Proto & 
	$63.69~|~64.89 ~(+6.41)$ & $69.64~|~69.74 ~(+8.29)$ & $70.42~|~70.52 ~(+4.48)$ &
	$51.12~|~51.51 ~(+5.71)$ & $53.80~|~54.01 ~(+3.34)$ & $57.89~|~58.28 ~(+3.79)$ \\

	SAT-PN & Self-Attn & Proto & 
	$68.09~|~68.79 ~(+10.31)$ & $71.03~|~71.25 ~(+9.80)$ & $72.33~|~72.47 ~(+6.43)$ &
	$58.09~|~58.42 ~(+12.62)$ & $60.43~|~61.57 ~(+10.90)$ & $65.01~|~65.89 ~(+11.40)$ \\

	DMN-PN & DMN & Proto & 
	$72.08~|~72.43 ~(+\underline{13.95})$ & $72.47~|~73.38 ~(+\underline{11.93})$ & 
	$73.91~|~74.68 ~(+\underline{8.64})$ &
	$59.95~|~60.07 ~(+\underline{14.27})$ & $61.48~|~62.13 ~(+\underline{11.46})$ & 
	$65.84~|~66.31 ~(+\underline{11.82})$ \\

	\midrule

	BRN-MPN & Bi-LSTM & M-Proto & 
	$63.19~|~63.78 ~(+\underline{5.30})$ & $65.16~|~65.33 ~(+\underline{3.88})$ & 
	$69.43~|~69.91 ~(+\underline{3.87})$ &
	$55.13~|~55.28 ~(+\underline{9.48})$ & $56.69~|~57.52 ~(+\underline{6.85})$ & 
	$61.25~|~61.76 ~(+\underline{7.27})$ \\

	CNN-MPN & CNN & M-Proto & 
	$66.01~|~66.87 ~(+\underline{8.39})$ & $68.06~|~68.17 ~(+\underline{6.72})$ & 
	$71.38~|~71.99 ~(+\underline{5.95})$ &
	$53.01~|~53.38 ~(+\underline{7.58})$ & $55.63~|~55.78 ~(+\underline{5.11})$ & 
	$59.34~|~60.08 ~(+\underline{5.59})$ \\ 

	SAT-MPN & Self-Attn & M-Proto & 
	$70.97~|~71.58 ~(+\underline{13.10})$ & $72.21~|~72.49 ~(+\underline{11.04})$ & 
	$73.64~|~74.12 ~(+\underline{8.08})$ &
	$60.10~|~60.55 ~(+\underline{14.75})$ & $62.45~|~62.82 ~(+\underline{12.15})$ & 
	$66.83~|~66.99 ~(+\underline{12.50})$ \\

	\textbf{DMB-PN} & \textbf{DMN} & \textbf{M-Proto} &
	$\bm{73.59}~|~\bm{73.86} ~(+\underline{\bm{15.38}})$ & $\bm{73.99}~|~\bm{74.82} ~(+\underline{\bm{13.37}})$ & $\bm{76.03}|~\bm{76.57} ~(+\underline{\bm{10.53}})$ & 
	$\bm{60.98}~|~\bm{62.44} ~(+\underline{\bm{16.64}})$ & $\bm{63.69}~|~\bm{64.43} ~(+\underline{\bm{13.76}})$ & $\bm{67.84}~|~\bm{68.35} ~(+\underline{\bm{13.86}})$ \\ 

	\bottomrule
	\end{tabular}
}
	\caption{Accuracy ~($\%$) and F1~Score $(10^{-2})$ of few-shot event classification.
	‘‘Encoder'' and ‘‘Metric'' denote the sentence encoder and the metric-based model respectively, so the final ‘‘Model'' is a combination of them.
	‘‘Match'', ‘‘Proto'' and ‘‘M-Proto'' are an abbreviation for matching network, prototypical network and memory-based prototypical network respectively. 
	The value$(+m) ~($\%$)$ in the brackets denotes the accuracy margin  calculated by subtracting the accuracy of the worst baseline from that of the current model under inspection.
	\label{tab:exp_effect_dmb_model} }
	
\vspace{-7mm}
\end{table*}

%% file: conclusion.tex
\section{Conclusion} \label{sec:con_fuw}
In this paper, we propose a dynamic-memory-based prototypical network (DMB-PN) for few-shot event detection task in meta-learning setting.
Our approach consists of two stages: trigger identification and few-shot event classification.
In the first stage, we locate the trigger in each event mention, and obtain memory-augmented sentence encoding based on DMN.
In the second stage, we utilize the dynamic-memory-based prototypical network to classify the event type of each query instance, where event mentions are encoded by utilizing the multi-hop mechanism of DMN to capture contextual information among event mention encodings.
The experiment results demonstrate that the integration of prototypical network and  dynamic-memory-based model excels at addressing the sample-shortage problem for few-shot event detection and dynamic-memory-based approaches are more feasible than other  sentence encoding baselines in context of limited labeled sentence instances, especially when the variety of event types is large and the instance quantity is small.

In the future, we will apply DMB-PN to other few-shot tasks, such as few-show relation extraction, trying to exploit the contexts of texts and entities.